\documentclass{article}

\usepackage{spconf}
\usepackage{amsmath}
\usepackage{amsfonts}
\usepackage{xcolor}
\usepackage{booktabs}
\usepackage{appendix}
\usepackage{newfloat}
\usepackage{hyperref}
\usepackage{algorithm}
\usepackage[noend]{algpseudocode}
\usepackage{multirow}
\usepackage{graphicx}
\usepackage{subcaption}
\usepackage{mwe}

\DeclareMathOperator*{\argmax}{arg\,max}

\algnewcommand{\LeftComment}[1]{\Statex \(\triangleright\) #1}

\title{Improved Activation Clipping for Universal
Backdoor Mitigation and Test-Time Detection}

\name{
\begin{tabular}{@{}c@{}}
Hang Wang\textsuperscript{\rm 1}, Zhen Xiang\textsuperscript{\rm 2}, David J. Miller\textsuperscript{\rm 1}, 
    George Kesidis\textsuperscript{\rm 1}
\end{tabular}
}
\address{
\begin{tabular}{ccc}
\textsuperscript{\rm 1} Anomalee Inc. \& Pennsylvania State University\\ 
\textsuperscript{\rm 2}University of Illinois at Urbana-Champaign\\
\end{tabular}
}

\begin{document}
%
\maketitle
\begin{abstract}
Deep neural networks are vulnerable to backdoor attacks (Trojans), where an attacker poisons the training set with backdoor triggers so that the neural network learns to classify test-time triggers to the attacker's designated target class. Recent work shows that backdoor poisoning induces overfitting (abnormally large activations) in the attacked model, which motivates a general, post-training
clipping method
for backdoor mitigation, {\it i.e.}
with bounds on
internal-layer activations learned using a small set of clean samples. We devise a new such approach, 
choosing the activation bounds to explicitly limit classification margins.  
This method gives superior performance against peer methods for CIFAR-10 image classification.
We also show that this method has strong robustness against adaptive attacks, 
X2X attacks, and on different datasets.
Finally, we
demonstrate a method extension for test-time detection and correction 
based on the output differences between the original and activation-bounded networks. \href{https://github.com/wanghangpsu/MMAC}{The code of our method is online available.} 
\end{abstract}

\section{Introduction} \label{sec:mitigation}

Deep Neural Networks (DNNs) have been 
successfully applied to various domains. However, a large amount of labeled data and extensive computational resources are required to train accurate DNNs. 
Thus, many users may 
employ 
untrusted third-party data, deep learning services, or even fully trained DNN models, all of which may be compromised by backdoor attacks.
A backdoor attacker \cite{BadNet} 
embeds a backdoor mapping into a DNN classifier by poisoning the training set or via malicious control of the training process.  The DNN will then classify test samples with the backdoor trigger to the attacker's ``target'' class. 
An effective backdoor attack requires little poisoning, uses a subtle (not easily detected) backdoor pattern, and has little to no effect on the DNN's accuracy on clean (trigger-free) test patterns.  

In this work, we consider image classification and focus on the post-training defense scenario. Here the defender is a downstream user of the trained classifier, without any access to the training set.  
The defender aims to ``repair'' the classifier so that i) misclassifications are not induced by backdoor trigger samples; and ii) good accuracy on benign (clean) examples is preserved.  Moreover,  the defender does not know whether an attack is present -- thus, mitigation is performed (and ii) should be achieved) even in the absence of backdoor poisoning.
The defender is assumed to have access to the DNN classifier and to a small set of benign samples, but no access to the training set or to any unattacked models. 

Most existing works on backdoor mitigation either fine-tune \cite{NC, I-BAU, NAD} or prune \cite{FP} the DNN classifier. NC \cite{NC} and I-BAU \cite{I-BAU} fine-tune the classifier using reverse-engineered backdoor triggers. Both implicitly assume some knowledge of how the backdoor triggers used to attack the model are embedded into source-class samples, and may not generalize well to models attacked by a different (particularly an unknown) type of backdoor trigger. Other fine-tuning methods (like NAD \cite{NAD}) make no assumption on the trigger type and fine-tune based only on benign samples. But they require a large number of such benign samples to remove the backdoor while also maintaining good clean (trigger-free) test set accuracy. Among pruning-based methods, Fine-Pruning \cite{FP} mitigates the backdoor by removing neurons deemed unnecessary for accurate classification of clean samples. However, some neurons may contribute to decisionmaking for both clean samples and backdoor triggers. 
Removing them may remove the backdoor mapping, but with significant degradation in clean test accuracy. Recognizing that backdoor triggers induce unusually large internal signals (and unusually large decisionmaking confidence) in the DNN, MMBM \cite{MMBD} imposes activation clipping (AC) on ReLU activations to defeat the backdoor.
However, this approach was ineffective against global trigger patterns \cite{Haoti} (which may make subtle changes even to {\it every} pixel in an image).

\begin{figure*}[t]
	\centering
	\includegraphics[width=1.0\linewidth]{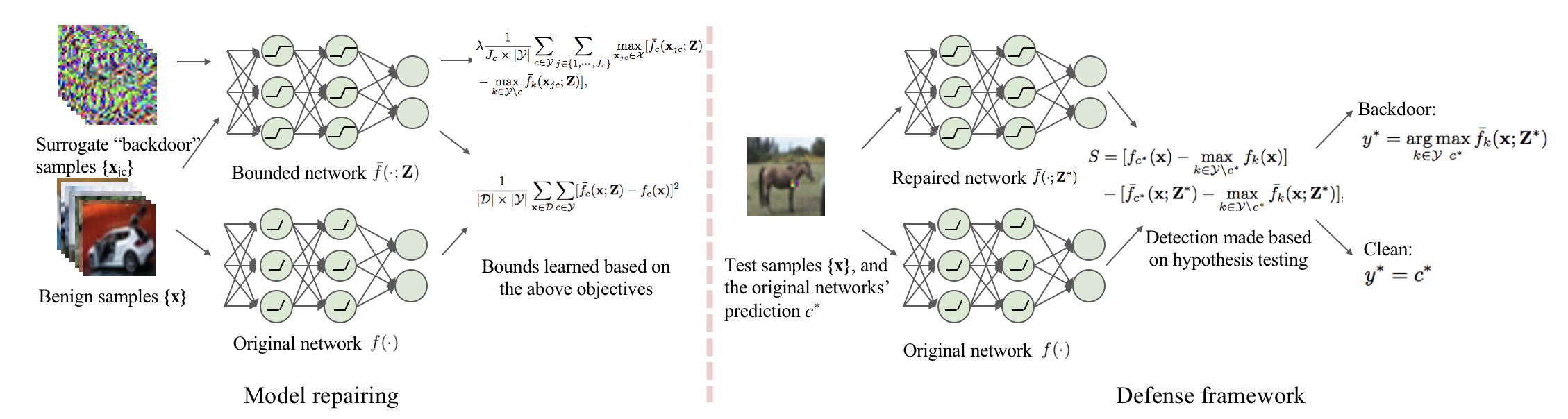} %
	\caption{Overview of our method.}%
\label{fig:overview}%
\end{figure*}

\cite{MMBD} did recognize that backdoors are fundamentally an overfitting phenomenon, with the experiments in \cite{MMBD} demonstrating that backdoors induce unusually large classification margins (decisionmaking confidence),
such that the backdoor pattern can overcome the features in the image that are representative of the true (source) class of origin.  Such overfitting thus ensures that backdoor triggers are classified to the attacker's designated target class, not to the source class of the sample.  However, \cite{MMBD} did not exploit classifier margin as a criterion for bounding neural activations -- it simply directly penalized against unusually large activations.
Here, instead, we choose bounds on activations to explicitly limit
the 
maximum classification margin (MM), assessed over the feasible input space of the DNN. 
Our mitigation approaches do not alter any DNN parameter values, {\it i.e.} they are not fine-tuning methods.  They are also not reverse engineering
based defenses, as they do not make any assumptions about the backdoor pattern or how it is incorporated into images ({\it e.g.}, additively or via patch replacement).  In fact, they can be applied without any knowledge of whether the DNN was backdoor-poisoned -- if poisoned, they are effective at mitigating a variety of backdoor attacks; if unpoisoned, they induce only modest degradation in clean test accuracy.
Furthermore, we experimentally show that our methods are robust against
adaptive attacks that exploit full knowledge of the defense.

\section{Background}

\subsection{Backdoor Attacks}

Backdoor attacks can be achieved simply by poisoning a small subset of the training set with backdoor triggers \cite{BadNet, Targeted, Haoti, nguyen2021wanet, Clean_Label_BA}.  The resulting trained DNN will misclassify test samples with backdoor triggers but behave normally on benign samples. \cite{BadNet} uses a replaced 
patch as the backdoor trigger. To increase stealthiness, some works use imperceptible backdoor triggers, {\it e.g.} \cite{Haoti} adds the backdoor trigger as an image perturbation, while \cite{Targeted} uses a ``blended'' pattern. \cite{nguyen2021wanet} adopts image wrapping as the backdoor trigger.  While these methods all apply mislabeling to the backdoor-poisoned training samples (labeling them to the attacker's designated target class), the clean-label attack in \cite{Clean_Label_BA} does not require any mislabeling. 
If the attacker has some control over the training process, 
an ``input-aware'' attack can be devised, wherein there is no common trigger, but rather a distinct trigger for each distinct input pattern\cite{nguyen2020inputaware}.
Given the rich space of existing (and possible) backdoor attacks, it is challenging to develop defense methods which can perform well on all of them ({\it i.e.}, achieving robust, universal performance).

\subsection{Post-Training Backdoor Defenses}

We
consider the post-training scenario where
 the defender (e.g., a downstream user) has no control of the training process and does not possess the training set. Some works proposed to detect whether a model is backdoor attacked and the associated target class(es) 
\cite{NC, Tabor, ABS, META, DataLimited,Post-TNNLS,L-RED}. 
Other works attempt to repair the potentially attacked model 
 as discussed in Sec. \ref{sec:mitigation}.
 Lastly, defenses can also be applied to detect whether a test sample is a backdoor trigger and, if so, attempt to correct the class decision, e.g., \cite{STRIP, InFlight}. 

\section{Improved Activation Clipping: MMAC}

Since they are widely applied, we assume here that the DNN uses ReLU activation functions, which are lower-bounded but positively unbounded. Thus, we can clip activations simply by applying upper bounds (saturation levels) to ReLUs.
Our method also works on DNNs with other activation functions, see 
Sec. \ref{sec:leaky_relu}.
We assume the defender has access to the trained DNN classifier $f$ and to a small clean dataset, $\mathcal{D}$. Let $f_c({\bf x})$ be class $c$'s logit output for sample ${\bf x}$ and let $h_l(\cdot)$ be the vector of outputs from the $l$-th (feedforward) layer. The logit function of an $L$-layer network can then be expressed as:
\begin{equation}
f_c({\bf x}) = h_L  (\cdots h_1 ({\bf x})),
\end{equation}
where we omit dependence on the DNN parameters since our method does not modify any parameter values. We repair the DNN by introducing a set of bounding vectors ${\bf z}_l$, $l=1, ..., L-1$, where $z_{lj}$ is the bound on the $j$-th ReLU in layer $l$.  Letting $\bar{h}_l(\cdot;z_l) = \min\{h_l(\cdot), {\bf z}_l\}$ be the bounded activations, we can express the bounded model as:
\begin{equation}
\bar{f}_c({\bf x;Z}) = h_L  (\bar{h}_{L-1}(\cdots \bar{h}_1 ({\bf x}; {\bf z}_1) \dots;{\bf z}_{L-1})), \forall c
\end{equation}
where ${\bf Z}=\{{\bf z}_1, \cdots, {\bf z}_{L-1}\}$. Notably, for convolutional layers, which produce spatially invariant feature maps, we apply the same bound to every neuron in a given feature map.  We choose the bounding vectors to minimize the following objective:
\begin{equation}\label{eq:mitigation}
\begin{split}
L({\bf Z}, \lambda; {\mathcal D}_{s})& = \frac{1}{|{\mathcal D}_{s}|\times|{\mathcal Y}|} \sum_{{\bf x}\in{\mathcal D}_{s}} \sum_{c\in{\mathcal Y}} [\bar{f}_c({\bf x};{\bf Z}) - f_c({\bf x})]^2 \\&+ \lambda \frac{1}{J_c \times |{\mathcal Y}|} \sum_{c\in{\mathcal Y}} \sum_{j \in \{1,\cdots ,J_c\}} \max_{{\bf x}_{jc} \in \mathcal{X}}  [\bar{f}_c({\bf x}_{jc};{\bf Z})
\\& 
\hspace{1in} - \max_{k\in \mathcal{Y}\setminus c}\bar{f}_k({\bf x}_{jc};{\bf Z})],
\end{split}
\end{equation}
where ${\mathcal D}_{s}$ represents the subset of samples in ${\mathcal D}$ from class $s$ correctly classified by the (unrepaired) classifier,
$\mathcal{Y}$ is the set of class labels, and $\lambda$ controls the weight given to the second term. 
The first term is the mean squared error (MSE) between the {\bf logit} outputs of the bounded network and the original network when benign samples are fed in. 
This term is based on the premise that the attack does not largely impact class logits on clean (benign) samples -- in such case, the logits provide more {\it informative} (real-valued) supervising targets than mere (discrete-valued) class labels.  This term helps to ensure that bounding will not severely degrade clean test set accuracy.
In the second term, for each class $c$, since margin maximization is a non-convex objective, we generate 
$J_c$ 
margin maxima
($\{{\bf x}_{jc}; j = 1, \cdots, J_c \forall c\}$) and limit the sum, over all classes, of the largest maxima.  To minimize \eqref{eq:mitigation}, we alternate gradient-based margin maximizations, yielding $\{{\bf x}_{jc}; j = 1, \cdots, J_c\}$, with gradient-based minimization of \eqref{eq:mitigation}, yielding {\bf Z}.
Moreover, we dynamically 
adjust $\lambda$ during the learning process to make sure the classification accuracy of the 
bounded model $\bar{f}$ on the benign samples is larger than a 
threshold $\pi$ (e.g. $\pi = 0.95$).
Finally, in all our experiments, we initialize all upper bounds in ${\bf Z}$ to 1 (an alternative is to initialize these bounds to the largest activation values induced by benign samples). 
By solving the optimization problem defined by Eq. \eqref{eq:mitigation}, we obtain the bounds ${\bf Z^*}$ and thus get a repaired DNN model $\bar{f}(\cdot; {\bf Z^*})$.
Note that, unlike reverse-engineering based mitigation methods \cite{NC, I-BAU} --
which rely 
on a previously applied detector to determine whether the DNN was attacked, to determine the target class, and to estimate the backdoor pattern -- our method does not rely on estimated backdoor patterns and an estimated target class. The margin maximizations are performed starting from randomly initialized points in the feasible input space, e.g., ${\mathcal X}=[0, 1]^{H\times W\times C}$ for color images with height $H$, width $W$, and $C$ channels, and the second term in \eqref{eq:mitigation} involves a sum over {\it all} classes. 
In this sense, our mitigator is ``universal'' -- it does not rely on any assumptions about the backdoor pattern type or its method of incorporation into images.  Moreover, it does not rely on the backdoor attack being successfully detected -- it is applied irrespective of whether the DNN has been backdoor-attacked.  In the sequel, we demonstrate our approach is successful at mitigating a wide variety of backdoor attacks, while only modestly degrading classification accuracy when there is no attack. 
The details on how to adjust 
$\lambda$ can be found in Apdx. \ref{apdx:mmac-details}.

\section{Improved Mitigation and Test-Time Detection by Hypothesis Testing} 
\label{sec:detection}
MMAC can be used to infer whether a test sample contains a backdoor trigger by comparing the class decisions produced by the original and the repaired DNNs.
If these decisions disagree, one may infer that the sample is a backdoor trigger.
Moreover, in this case, one can estimate the true class as the decision produced by the repaired network.  However, even if the two class decisions agree, the test sample could be a backdoor trigger -- {\it e.g.}, the test sample, from class $s \neq t$, could be misclassified as class $t$ by the original network (especially if there is large confusion between classes $s$ and $t$).  In this section, we extend MMAC to improve test-time detections and classifications, with the resulting method referred to as MM Defense Framework (MMDF).

Given any test sample ${\bf x}$ and its class decision $c^*$ produced by the original network, a detection statistic can be calculated:
\begin{equation}
\label{eq:stat}
\begin{split}
S &= [f_{c^*}({\bf x}) - \max_{k\in \mathcal{Y}\setminus c^*}f_k({\bf x})] 
\\& - [\bar{f}_{c^*}({\bf x};{\bf Z}^*) - \max_{k\in \mathcal{Y}\setminus c^*}\bar{f}_k({\bf x};{\bf Z}^*)],
\end{split}
\end{equation}
which is the margin change between the original and MMAC-repaired networks. 
We hypothesize that bounding the network will reduce the margin of samples with the backdoor trigger while keeping the margin of benign samples unchanged. Thus,
when there is a backdoor trigger, even when the class decisions between the original and bounded networks agree, the statistic 
in (\ref{eq:stat}) will be abnormally large. 
See Apdx. \ref{apdx:dist}.
Thus we can estimate a null distribution $H_0$ (e.g., Gaussian distribution)
for this detection statistic
using all the benign samples in the clean set $\mathcal{D}$. And we declare a detection when the p-value is less than $\theta = 0.005$, {\it i.e.} with a 99.5\% confidence level. Notably under this threshold setting, we allow theoretically a 0.5\% false positive rate. But since there is usually only a small clean set available such that one cannot learn very accurate null distributions, the empirical false positive rate may differ from 0.5\%.

If a test sample is detected as benign, the user chooses the classification decision of the original network. Whereas for a sample identified as attacked, the defender either 
rejects the sample (make no decision)
or chooses the class decision made by the bounded network, {\it excluding} the original network's predicted class $c^*$. 

Both the MMAC and MMDF approaches are illustrated in Figure 1.
\begin{table*}[!t]
	
	\begin{center}
        \scriptsize
	\begin{tabular}{ccccccccccc}
				\toprule
				
				&$N_{\rm img}$ &  & chessboard   & BadNet  & blend  & WaNet  &Input-aware &Clean-Label&no attack&
				\\
				\midrule
				\multirow{3}{5.5em}{without defense}&\multirow{3}{0.5em}{0} &PACC&0.59$\pm$0.84&0.57$\pm$0.76&4.67$\pm$7.45&3.48$\pm$1.83&5.50$\pm$3.30&5.72$\pm$5.36&n/a\\	
    &&ASR&99.26$\pm$1.14&99.41$\pm$0.76&95.08$\pm$7.79&96.25$\pm$1.94&94.26$\pm$3.45&94.13$\pm$5.56&n/a\\
    &&ACC&91.26$\pm$0.35&91.63$\pm$0.35&91.57$\pm$0.39&90.94$\pm$0.30&90.65$\pm$0.34&88.68$\pm$1.26&91.41$\pm$0.16\\
				\cline{3-11}
    \multirow{3}{5.5em}{NC}&\multirow{3}{0.5em}{500} &PACC&31.98$\pm$24.44&62.03$\pm$17.80&25.84$\pm$33.09&0.30$\pm$0.38&11.24$\pm$9.38&83.85$\pm$1.81&n/a\\	
    &&ASR&63.72$\pm$27.20&28.13$\pm$19.43&71.01$\pm$36.94&99.64$\pm$0.45&86.10$\pm$11.73&20.72$\pm$13.84&n/a\\
    &&ACC&88.49$\pm$1.66&87.86$\pm$1.70&87.10$\pm$2.59&72.25$\pm$10.60&81.81$\pm$1.25&68.37$\pm$10.04&n/a\\
				\cline{3-11}
    \multirow{3}{5.5em}{NAD}&\multirow{3}{0.5em}{200} &PACC&79.88$\pm$1.92&81.31$\pm$3.47&86.43$\pm$0.63&65.33$\pm$12.62&72.07$\pm$4.24&76.11$\pm$5.83&n/a\\	
    &&ASR&2.92$\pm$1.56&5.52$\pm$2.94&1.92$\pm$0.70&18.03$\pm$14.46&5.96$\pm$7.37&10.54$\pm$9.17&n/a\\
    &&ACC&81.28$\pm$1.75&86.52$\pm$1.53&87.07$\pm$1.30&82.39$\pm$8.81&79.92$\pm$4.02&82.76$\pm$1.39&88.87$\pm$0.51\\
    \cline{3-11}
    \multirow{3}{5.5em}{Fine-Pruning}&\multirow{3}{0.5em}{500} &PACC&19.25$\pm$16.53&10.91$\pm$14.41&14.47$\pm$13.19&83.03$\pm$6.40&86.04$\pm$1.96&74.60$\pm$8.91&n/a\\	
    &&ASR&55.53$\pm$37.34&86.49$\pm$17.35&84.44$\pm$14.02&0.37$\pm$0.23&4.09$\pm$2.71&13.81$\pm$11.50&n/a\\
    &&ACC&90.73$\pm$0.60&91.83$\pm$0.77&92.78$\pm$0.66&90.71$\pm$0.78&90.71$\pm$1.17&84.57$\pm$1.42&90.61$\pm$0.77\\
    \cline{3-11}
    \multirow{3}{5.5em}{I-BAU}&\multirow{3}{0.5em}{100} &PACC&84.59$\pm$3.97&85.49$\pm$3.32&82.45$\pm$8.53&81.95$\pm$5.63&80.59$\pm$4.27&56.95$\pm$18.51&n/a\\	
    &&ASR&3.16$\pm$3.10&2.45$\pm$3.74&6.61$\pm$8.78&4.62$\pm$5.64&1.62$\pm$1.06&35.59$\pm$21.93&n/a\\
    &&ACC&89.62$\pm$0.45&89.74$\pm$0.57&87.87$\pm$1.11&87.93$\pm$2.40&89.37$\pm$0.71&86.69$\pm$1.85&89.30$\pm$0.43\\
    \cline{3-11}
    \multirow{3}{5.5em}{MMBM}&\multirow{3}{0.5em}{50} &PACC&14.19$\pm$21.55&86.53$\pm$1.19&86.92$\pm$3.96&78.23$\pm$9.35&79.75$\pm$1.47&85.79$\pm$1.56&n/a\\	
    &&ASR&81.77$\pm$25.58&1.49$\pm$0.87&3.45$\pm$1.73&9.85$\pm$11.33&1.57$\pm$1.16&1.79$\pm$0.9&n/a\\
    &&ACC&88.48$\pm$0.61&89.12$\pm$0.77&88.67$\pm$0.36&86.24$\pm$9.36&86.08$\pm$1.41&86.16$\pm$1.10&88.70$\pm$0.44\\
				\cline{3-11}
    \multirow{3}{5.5em}{MMAC (ours)}&\multirow{3}{0.5em}{50} &PACC& 52.12$\pm$32.08&83.57$\pm$3.06&85.18$\pm$2.64&84.20$\pm$1.75&81.86$\pm$2.23&85.76$\pm$1.26&n/a\\	
    &&ASR& 26.19$\pm$36.49&2.21$\pm$1.84&$4.00\pm$2.51&4.24$\pm$2.32&1.70$\pm$1.04&1.74$\pm$0.99&n/a\\
    &&ACC&87.92 $\pm$0.97&88.42$\pm$0.79&88.23$\pm$0.68&88.00$\pm$0.95&88.98$\pm$0.64&86.21$\pm$1.16&87.47$\pm$0.94\\
    			\cline{3-11}
    \multirow{3}{5.5em}{MMDF (ours)}&\multirow{3}{0.5em}{50} &PACC& 67.35$\pm$15.72&86.01$\pm$1.50&86.92$\pm$1.63&85.02$\pm$2.37&80.81$\pm$2.53&86.81$\pm$1.63&n/a\\	
    &&ASR&0.45$\pm$0.64&0.09$\pm$0.14 &1.24$\pm$1.18&2.66$\pm$1.31&3.37$\pm$2.40&0.01$\pm$0.01&n/a\\
&&ACC&89.12$\pm$0.44&89.42$\pm$0.41&90.04$\pm$0.40&87.02$\pm$0.65&89.22$\pm$0.44&86.79$\pm$1.41&88.99$\pm$0.45\\
    \bottomrule
		\end{tabular}	
  \caption{Performance of different mitigation methods under various attacks on CIFAR-10
  classifer.} 
		\label{tab:mitigation}
	\end{center}
\end{table*}

\section{Experiments}
\subsection{Setup}

{\bf Dataset:} experiments were conducted on CIFAR-10 \cite{CIFAR10}, with 60000 32$\times$32 color images from 10 classes. For each class, there are 5000 training images and 1000 test images. 
(See Sec. \ref{sec:other} for other datasets.)

{\bf Attack settings: } we consider the backdoor attacks from Sec. 2.1: 
1) The Chessboard pattern, a global additive pattern with intensity 3/255.
2) The BadNet attack, which uses a random 3 $\times$ 3 replaceable patch; the location of the patch is randomly selected and kept fixed for all poisoned training samples and all trigger test samples.
3) A local blended pattern, which uses the same pattern as BadNet, but blended into the image with a blend ratio of 0.2.  
4) The warping-based global pattern (WaNet).
5) A sample-specific attack utilizing a generator to produce a customized trigger for each input sample (Input-aware).
6) The clean-label attack, which does not mislabel samples during data poisoning. 
In all our experiments, we consider the all2one setting where one class is the target class and the remaining classes are source classes of the attack. Note that our defense does not make any assumption on the number of source or target classes-- each class is treated as a potential target class, with its maximum margin minimized/limited.
(See Sec. \ref{sec:x2x} for other source/target class settings.)

{\bf Performance criteria:} We evaluate performance using three criteria: ACC, ASR, and PACC. ACC is the classification accuracy on clean test samples, ASR (attack success rate) is the fraction of test samples embedded with backdoor triggers (backdoor samples) that are (mis)classified to the target class.
PACC (poisoned sample accuracy) is the percentage of test-time triggers correctly classified to the original (source) class. After mitigation, we want to maintain the utility of the repaired model, so it should have high ACC; for backdoor samples, the defender aims at least to not classify them to the target class (low ASR), but also to classify them to the correct source class (high PACC).

\subsection{Results on CIFAR-10}
In this section, MMAC and MMDF are compared with five previous methods: NC \cite{NC}, I-BAU \cite{I-BAU}, NAD \cite{NAD}, Fine-Pruning \cite{FP}, and MMBM \cite{MMBD}. NC, I-BAU, and NAD are fine-tuning based methods, where NC and I-BAU estimate backdoor triggers and use them to fine-tune the model so that the repaired model will not make misclassifications on backdoor triggers. NC assumes the trigger is a replaceable patch while I-BAU assumes the trigger is an additive perturbation. NAD does not rely on estimation of backdoor triggers. It first uses the available benign samples to fine-tune the original model, yielding a ``weak'' model; it then uses the ``weak'' model to ``teach'' the original model, aiming to eliminate the backdoor. Such fine-tuning requires a large amount of clean data to maintain ACC. Fine-Pruning and MMBM, like MMAC and MMDF, do not alter DNN parameters.   

Table \ref{tab:mitigation} shows the results of all the methods under the different attacks.
As a fine-tuning method, NC relies on correctly estimating the backdoor trigger to ``unlearn" the backdoor. 
However, NC may not reliably estimate the trigger because the attack may not
conform to NC's assumptions that: i) all classes other than the target class
are source classes of the attack (the all2one assumption); ii) the trigger is patch-incorporated into images.
Thus, even with knowledge of the target class, NC cannot achieve very low ASR on any of the attacks. On the other hand,  NAD fine-tuning achieves low ASR across all the attacks; but with limited samples available for fine-tuning, there are severe drops in ACC. I-BAU has the best performance among fine-tuning-based methods. Instead of estimating a universal perturbation to induce misclassifications, I-BAU learns a separate perturbation for every image, and then fine-tunes the model so that these perturbations no longer cause misclassifications. Note that fine-tuning based methods usually require more clean samples (than alternative methods) to achieve good ACC. 

\begin{figure}[t]
	\centering	\subfloat[\centering Chessboard. ]{{\includegraphics[width=4cm]{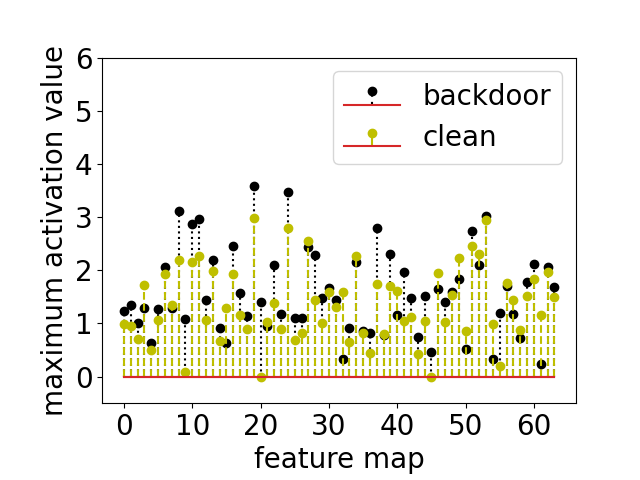}} }\label{fig:adaptive_asr}%
	~
	\subfloat[\centering BadNet. ]{{\includegraphics[width=4cm]{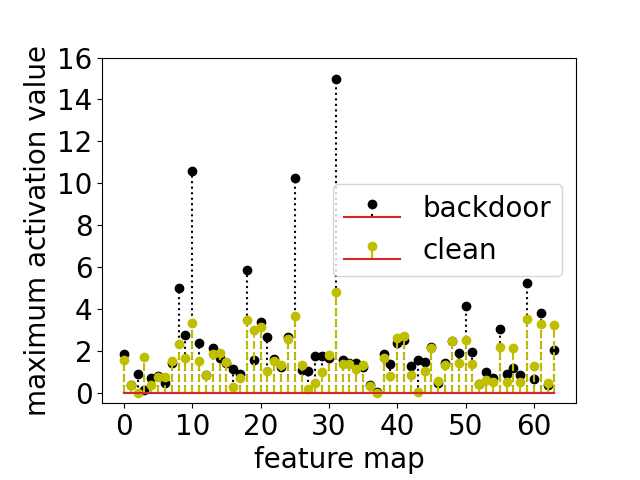} }}%
	\caption{Maximum activation values of each feature map for backdoor triggers and for clean samples, under (a) chessboard attack \cite{Haoti}  and (b) BadNet attack \cite{BadNet}. Note that the activation values for backdoor triggers are more greatly inflated under the Badnet attack than under the global chessboard attack.}%
 \label{fig:activations}
\end{figure}

Fine-Pruning has good performance on WaNet and Input-aware attacks. But as shown in Figure \ref{fig:activations}, some neurons (feature maps) are shared by benign samples and backdoor triggers. 
This is indicated by neurons that produce large activations for both clean samples and for backdoor triggers.
Dropping out neurons (or feature maps) that are not used for classifying clean samples might not eliminate the backdoor. So, the performance of Fine-Pruning on the other attacks is poor.  MMBM has good overall performance, except on the chessboard attack. As shown in Fig. \ref{fig:activations}, the internal activations caused by the chessboard pattern are just slightly larger than those induced by clean samples, so aggressively upper-bounding all the activations might not effectively mitigate the chessboard backdoor. 
MMAC has better performance on the chessboard, but only reduces the average ASR to 26.19\%, and with large error bars.
MMDF 
gives strong performance on the chessboard pattern and the best overall performance.
Note also that both MMAC and MMDF have good ACC, with only modest degradation compared with the ``no attack'' ACC of 91.41\%.  Finally, note that, in the last column, we report the ACC when mitigation is performed in the absence of an attack.  Note that all the methods yield good ACCs, with only modest degradation compared to the clean classifier's ACC of 91.41\%.

\subsection{Other activation functions}
\label{sec:leaky_relu}
\begin{figure}[ht]
	\centering
	\subfloat[\centering ]{{\includegraphics[width=4cm]{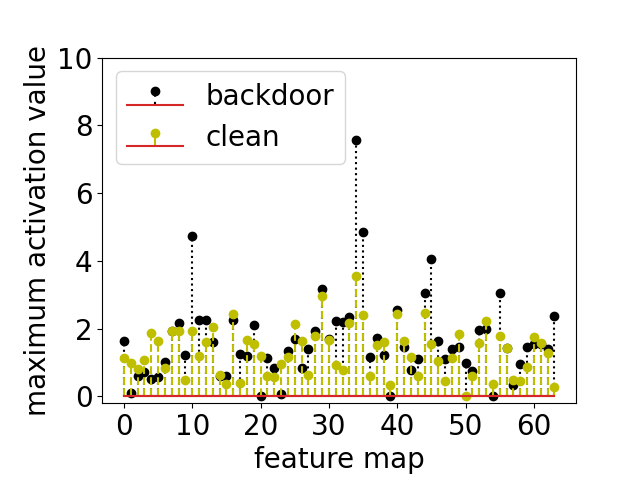}} }%
	~
	\subfloat[\centering ]{{\includegraphics[width=4cm]{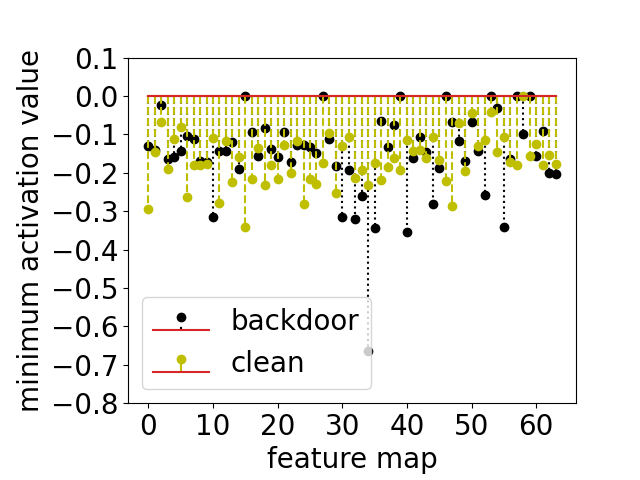} }}%
	\caption{Minimum \& maximum LeakyReLU activations.}%
\label{fig:leaky_relu}
\end{figure}

Since most DNN classifiers use ReLU activation functions, the method is proposed based on the ReLU activation and only upper bounds are introduced. In this section, we consider completely unbounded activation functions (e.g., LeakyReLU).  For such functions, we need to apply both upper-bounds and lower-bounds to activations. Here we consider a ResNet-18 DNN trained on the CIFAR-10 dataset that is poisoned with a BadNet attack. In the DNN, all the ReLU activations are replaced by LeakyReLUs, with a negative slope of 0.1. Fig. \ref{fig:leaky_relu} shows the minimum and maximum activation values in each feature map produced by the first convolutional layer (after the activation function is applied). The plots indicate that backdoor triggers induce large absolute internal layer activations; thus, the model might be repaired by applying both lower bounds and upper bounds to the activation values. As in our main experiments, the bounds can be learned using the objective defined in \eqref{eq:mitigation}. The performance is reported in Table \ref{tab:leaky_relu}, which shows our methods achieve low ASRs while maintaining the ACC.  This further demonstrates that our methods
are applicable to models without ReLU activations.
\begin{table}[ht!]
	\scriptsize
	\begin{center}
		\begin{tabular}{cccccccc}
			\toprule
                &No defense& MMAC& MMDF\\
                \hline 
               PACC&0.02&75.73&78.44\\
                ASR&99.98&4.77&0.01\\
                ACC&91.96&87.16&90.07\\
			\bottomrule
		\end{tabular}
		\caption{Defense performance against BadNet on ResNet-18 model with LeakyReLU activation function.} 
		\label{tab:leaky_relu}
	\end{center}
\end{table}

\subsection{Robustness against an adaptive attack}
To make a successful attack even when our defense is in play, the attacker needs to ensure that the activations caused by samples with a backdoor trigger are not larger (or not much larger) than the activations of benign samples. Otherwise, backdoor trigger samples may be detected by our MMDF method. Here we experimentally evaluate the robustness of our method against an adaptive attack. We assume the attacker fully controls the training process and training data, has full knowledge about our defense method (mitigation and detection methods), and even has access to the small clean set available for mitigation. In an attempt to defeat our defense, the attacker can alternate i) estimation of the upper bounds based on benign samples and ii) fine-tuning the model parameters based on the poisoned dataset.  
The attacker can use the objective  \eqref{eq:mitigation} and the small clean set (available to the defender) to learn the upper bounds ${\bf Z}$. The model can be fine-tuned using the following objective:
\begin{equation}
\begin{split}
&L_{adap} = \frac{1}{|{\mathcal D}_p|}\sum_{({\bf x}, y)\in{\mathcal D}_p} L_{ce}(\bar{f}({\bf x};{\bf Z}), y) + \beta \frac{1}{L \times |{\mathcal D_b}|}\\& \times \sum_{({\bf x}, y)\in{\mathcal D_b}}\sum_{l=1}^{L-1} \frac{1}{M_l} \sum_{m=1}^{M_l}\max(0, \phi_{lm}({\bf x}) - {\bf z}_{lm}),
\end{split}
\end{equation}
where the first term is the cross entropy loss of the whole poisoned training set ${\mathcal D_p}$ (including both benign and backdoor trigger samples). The second term penalizes the backdoor samples (subset ${\mathcal D_b}$) with activations larger than the upper bounds. $\phi_{lm}({\bf x})$ is the $m$-th activation from the $l$-th layer's output for sample ${\bf x}$, and ${\bf z}_{ml}$ is the corresponding upper bound. $\beta$ is the relative weight given to the two terms. The upper bounds should be updated iteratively since, after fine-tuning the model, the activations of benign samples may change. 
As in our main experiments, we use the CIFAR-10 dataset and ResNet-18 network architecture. The BadNet pattern is used by the attacker. Since the adaptive attacker has full control of the training process, we assume that all the training samples can be poisoned. A backdoor sample is created from each benign training sample that is not from the target class. Both the benign training sample and its poisoned version are included in the training set. 
In Tab. \ref{tab:adaptive}, we report the ACC and ASR for different values of $\beta$. Since every sample is poisoned and the same backdoor trigger repeatedly occurs in every backdoor sample and thus is easier to learn, when there is no defense, the ASR is always high. However, with increased $\beta$, the adaptive attacker focuses more on the activations of backdoor samples, and the ACC under `no defense' becomes lower. An abnormally low ACC is usually considered an indication of an unsuccessful backdoor attack.
Also with the increase of $\beta$, the adaptive attacker starts to defeat our MMAC method (the ASR increases), but only with a significant drop in ACC. Even when $\beta=10$, the ASR is vey low when MMDF is applied. These results suggest that our method has strong robustness against adaptive attacks. 
\begin{table}[t!]
	\scriptsize
	\begin{center}
		\begin{tabular}{cccccccc}
			\toprule
                &$\beta$  &0.5& 1& 5& 10\\
                \midrule
                \multirow{2}{5.5em}{No defense}&ACC &90.03&88.68&82.33&77.77\\
                &ASR&100.00&100.00&100.00&100.00\\
                \cline{2-6}
                 \multirow{2}{5.5em}{MMAC}&ACC& 87.13&85.29&80.51&75.21\\
                &ASR&2.48&15.76&39.40&53.08\\
                \cline{2-6}
                 \multirow{2}{5.5em}{MM-defense}&ACC &89.03&86.83&80.62&76.85\\
                &ASR&0.13&0.30&5.57&2.28\\
			\bottomrule
		\end{tabular}
		\caption{Adaptive attack results for different $\beta$, when there is no defense, MMAC, or MM-defense framework.} 
		\label{tab:adaptive}
	\end{center}
\end{table}

\begin{table*}[!t]
	\scriptsize
	\begin{center}
 \setlength{\tabcolsep}{2.5pt}
			\begin{tabular}{cccccccccccccc}
				\toprule
				& && \multicolumn{3}{c}{CIFAR-100}& \multicolumn{2}{c}{TinyImageNet} &   \multicolumn{4}{c}{GTSRB}   \\ \cmidrule(lr){4-6} \cmidrule(lr){7-8}\cmidrule(lr){9-12}
				&$N_{img}$& & no attack & Chessboard  & BadNet & no attack & BadNet &no attack &chessboard&BadNet&blend
				\\
				\midrule
				\multirow{3}{5.5em}{without defense}&\multirow{3}{0.5em}{0} &PACC&n/a&2.16$\pm$1.21&0$\pm$0&n/a&1.71$\pm$0.47&n/a&0.11$\pm$0.08&0$\pm$0&0.68$\pm$0.35\\	
    &&ASR&n/a&97.16$\pm$1.76&100.00$\pm$0&n/a&96.95$\pm$1.07&n/a&99.90$\pm$0.06&100$\pm$0&99.27$\pm$0.37\\
    &&ACC&67.49$\pm$0.26&65.33$\pm$0.53&66.19$\pm$0.16&57.92$\pm$2.86&58.95$\pm$0.54&94.86$\pm$0.50&95.04$\pm$0.39&95.02$\pm$0.45&95.11$\pm$0.53\\
				\cline{3-12}
    \multirow{3}{5.5em}{MMAC}&\multirow{3}{0.5em}{5} &PACC&n/a&40.31$\pm$14.00&56.13$\pm$2.79&n/a&52.76$\pm$0.71&n/a&68.18$\pm$29.34&87.60$\pm$3.38&88.94$\pm$2.37\\	
    &&ASR&n/a&26.60$\pm$23.94&3.97$\pm$2.15&n/a&1.71$\pm$2.03&n/a&19.12$\pm$33.54&0.11$\pm$0.11&0.47$\pm$0.44\\
    &&ACC&66.38$\pm$0.55&61.75$\pm$0.40&65.65$\pm$0.12&57.42$\pm$2.92&58.38$\pm$0.49&90.08$\pm$1.62&90.49$\pm$0.57&91.38$\pm$0.59&90.87$\pm$1.19\\
				\cline{3-12}
    \multirow{3}{5.5em}{MMDF}&\multirow{3}{0.5em}{5} &PACC&n/a&46.70$\pm$8.48&57.29$\pm$2.38&n/a&53.14$\pm$1.40&n/a&68.78$\pm$27.56&87.63$\pm$3.35&88.71$\pm$2.58\\	
    &&ASR&n/a&6.85$\pm$7.95&0$\pm$0&n/a&0$\pm$0&n/a&17.95$\pm$31.03&0.0$\pm$0.0&0.87$\pm$1.59\\
    &&ACC&66.31$\pm$0.41&61.04$\pm$0.57&65.06$\pm$0.15&56.89$\pm$3.20&58.22$\pm$0.51&90.39$\pm$2.04&91.28$\pm$1.26&91.39$\pm$1.28&91.66$\pm$1.64\\
				\bottomrule		
		\end{tabular}
	\caption{Other datasets}
		\label{tab:other_datasets}
	\end{center}
\end{table*}

\subsection{Results on other datasets}
\label{sec:other}
Here we evaluate MMAC and MMDF on other datasets (CIFAR-100, TinyImagenet, and GTSRB). 
Due to the complex attack processes for WaNet, Input-aware, and Clean-label attacks, which make it difficult for these attacks to be successful on these datasets,  we do not include these attacks in the experiments in this section. 
For each type of attack, 5 models are trained and the average and the standard deviation of PACC, ASR, and ACC are reported in Table \ref{tab:other_datasets}.  The results show that our defense methods are effective in substantially reducing ASRs (the highest average ASR for MMDF is 17.95 \%, on GTSRB for the chessboard attack).

\begin{table*}[!t]
\scriptsize
	\begin{center}
 \setlength{\tabcolsep}{2.5pt}
			\begin{tabular}{cccccccccccccc}
				\toprule
				&& \multicolumn{2}{c}{chessboard}& \multicolumn{2}{c}{BadNet} &   \multicolumn{2}{c}{blend} \\ 
    \cmidrule(lr){3-4} \cmidrule(lr){5-6}\cmidrule(lr){7-8}
				&&one2one& all2all & one2one  & all2all
				&one2one&all2all\\
				\midrule
				\multirow{3}{4.5em}{without defense} &PACC&0.11$\pm$0.09&1.24$\pm$0.23&0.09$\pm$0.13&0.09$\pm$0.07&7.94$\pm$10.96&0.05$\pm$0.04\\	
    &ASR&99.70$\pm$0.17&89.64$\pm$0.77&99.88$\pm$0.19&99.73$\pm$0.23&90.84$\pm$12.31&99.11$\pm$1.30\\
    &ACC&91.19$\pm$0.24&91.88$\pm$0.26&91.10$\pm$1.64&91.75$\pm$0.47&91.10$\pm$0.99&91.60$\pm$0.50\\
    \cline{3-8}
    \multirow{3}{4.5em}{MMAC} &PACC&30.02$\pm$31.62&85.70$\pm$1.94&80.53$\pm$4.43&82.78$\pm$2.46&71.33$\pm$4.78&73.88$\pm$9.45\\	    &ASR&57.96$\pm$39.91&2.42$\pm$0.68&2.78$\pm$1.36&3.51$\pm$2.04&16.42$\pm$9.12&6.71$\pm$4.03\\
&ACC&88.02$\pm$1.10&87.38$\pm$1.48&86.73$\pm$2.12&87.92$\pm$0.69&88.03$\pm$0.83&87.54$\pm$0.86\\
    \cline{3-8}
    \multirow{3}{4.5em}{MMDF} &PACC&63.83$\pm$14.69&79.30$\pm$3.03&81.90$\pm$4.74&84.22$\pm$1.59&77.89$\pm$8.59&76.97$\pm$6.95\\
    &ASR&4.56$\pm$8.57&9.95$\pm$2.75&0.4$\pm$0.56&0.85$\pm$0.44&9.22$\pm$10.59&2.51$\pm$1.39\\
    &ACC&88.77$\pm$0.59&90.55$\pm$0.77&88.94$\pm$1.94&89.92$\pm$0.47&89.18$\pm$1.00&89.69$\pm$0.56\\
				\bottomrule		
		\end{tabular}
	\caption{X2X attacks}		\label{tab:x2x}
	\end{center}
\end{table*}

\subsection{X2X attacks}
\label{sec:x2x}
Until now
we have only considered all2one attacks, i.e., 
there is only one attack target class and all other classes are considered attack source classes. Note that our method does not make assumptions about the number of source or target classes, so it should be effective for arbitrary numbers of source and target classes. In this section, we consider two other scenarios: one2one and all2all attacks. In a one2one attack, there is one source class and one target class randomly selected from all the classes. In all2all attacks, samples with backdoor patterns originating from any class $c\in{\mathcal Y}$ will be mislabeled as class $(c+1)mod|{\mathcal Y}|$. We create attacked models using chessboard, BadNet, and blended attacks. For each X2X and backdoor trigger combination, we created 5 attacked models. The average and standard deviation of ASR, ACC, and PACC are reported in Tab. \ref{tab:x2x}. The results show that MMDF is generally effective under all scenarios.

\subsection{Limitations}
We now mention two limitations of our work. First, 
the MMAC objective defined in \eqref{eq:mitigation} requires generating magin maxima every several iterations, which is time-consuming. MMBM takes about 25 seconds to process one ResNet-18 model on CIFAR-10. MMAC, on the other hand, takes about 274 seconds to process one model. 
Second, note that MMAC and MMDF 
do not change any model parameters.  
It is possible that  
the PACCs for these methods could be improved, e.g., on the chessboard attack or the Input-aware attack, if these methods were combined with some type of model fine-tuning.

\section{Conclusions}

An activation clipping based backdoor mitigation approach was proposed that 
explicitly limits the maximum classification margins produced by the DNN.  
The resulting MMAC and MMDF methods,
the latter employing both the original and activation-clipped DNNs,
were shown to substantially mitigate backdoor attacks, while having only modest impact on the DNN's accuracy on
 a clean test set.

\section*{Acknowledgements}
This research was supported by NSF 
Grant No. 2132294.


\clearpage
\bibliographystyle{plain}

\clearpage
\appendix

\section{MMAC Algorithmic Details}\label{apdx:mmac-details}
\label{sec:algorithm}
~

\begin{algorithm}[h]
\vspace{0.1in}
        \small
	\caption{MMAC algorithm based on  Eq. \eqref{eq:mitigation}.}\label{alg:mitigation}
	\begin{algorithmic}[1]
		\State {\bf Inputs}: Small clean dataset ${\mathcal D}_s$, potentially attacked DNN model $f(\cdot)$, accuracy constraint $\pi$, learning rate $\delta$, maximum iterations $T_{max}$, number of iterations before performing more margin maximizations ($T$), and the scaling factor $\alpha$ used to adjust $\lambda$ in Eq. \eqref{eq:mitigation}.

		\State {\bf Initialization}: ${\bf Z}^{(0)}$ is initialized as small positive values (e.g. 1), $\lambda^{(0)}$ set to a small positive number (e.g. $10^{-1}$).
		\For{$l = 1:T_{max}$}
            \State{apply the bounds on $f(\cdot)$ to form the bounded network $\bar{f}(\cdot;{\bf Z})$}
            \If{$l \% T == 0$} \Comment{\textcolor{gray}{ generate surrogate samples}}
            \For{$c \in {\mathcal Y}$}
            \For{$j =1:J_c$} \Comment{\textcolor{gray}{ Generate $J_c$ samples for each class, which can be done in batches in practice}}
            \State{Randomly initialize $x_{jc} \in {\mathcal X}=[0, 1]^{H\times W\times C}$}
            \State \parbox[t]{170pt} {Optimize $x^*_{jc} = \argmax_{x_{jc}} [\bar{f}_c({\bf x}_{jc};{\bf Z}) - \max_{k\in \mathcal{Y}\setminus c}\bar{f}_k({\bf x}_{jc};{\bf Z})]$ using gradient ascent. }
            \EndFor
            \EndFor
            \EndIf
            
		\State ${\bf Z}^{(l+1)} = {\bf Z}^{(l)} - \delta \nabla_{\bf Z} L({\bf Z}^{(l)}, \lambda^{(l)}; {\mathcal D}_s)$
		\If{$\frac{1}{|{\mathcal D}|}\sum_{({\bf x}, y)\in{\mathcal D}}{\mathbf 1}[y=\argmax_{c\in{\mathcal Y}} \bar{f}_c({\bf x}; {\bf Z})] \geq \pi$}
		\State $\lambda^{(l+1)} = \lambda^{(l)}\cdot\alpha$
		\Else
		\State $\lambda^{(l+1)} = \lambda^{(l)}/\alpha$ \Comment{\textcolor{gray}{ Adjust $\lambda$ so that the classification accuracy of the bounded network $\bar{f}(\cdot; {\bf Z})$ on the small clean set ${\mathcal D}_s$ is larger than $\pi$ }}
		\EndIf
		\EndFor
		\State {\bf Outputs}: Optimized bounds ${\bf Z}^{\ast}$, and the bounded network $\bar{f}_c({\bf x}; {\bf Z})$
	\end{algorithmic}
\end{algorithm}

\section{The detection statistics of clean and backdoor samples and ROC curves}
\label{apdx:dist}

\begin{figure*}[ht]
	\centering
	\begin{minipage}[b]{.30\linewidth}
		\centering
		\centerline{\includegraphics[width=1.0\linewidth]{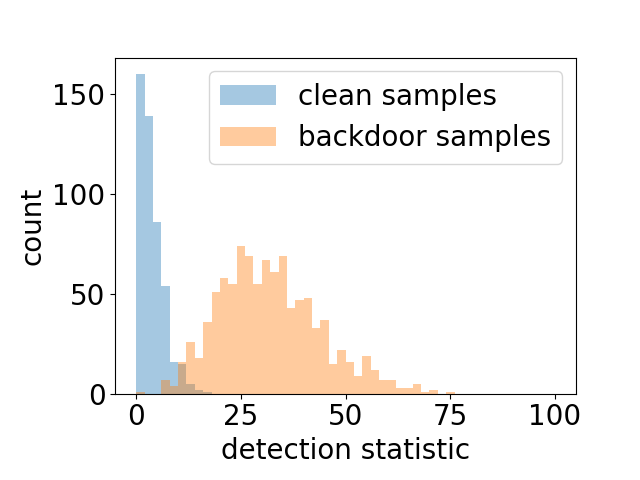}}
		\subcaption*{chessboard}
		\centerline{\includegraphics[width=1.0\linewidth]{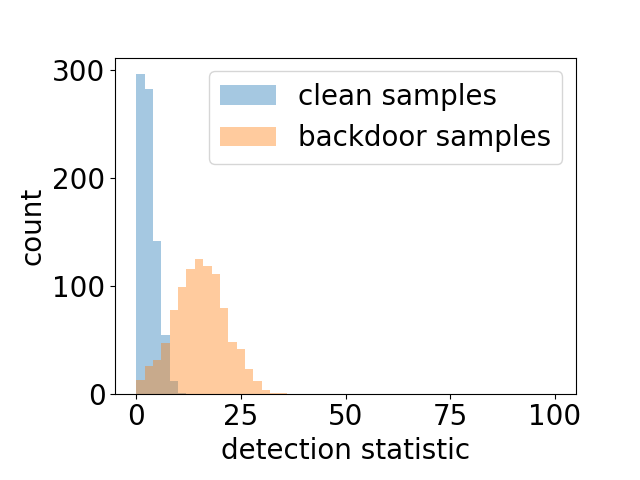}}
		\subcaption*{WaNet}
	\end{minipage}
	\begin{minipage}[b]{.30\linewidth}
		\centering
		\centerline{\includegraphics[width=1.0\linewidth]{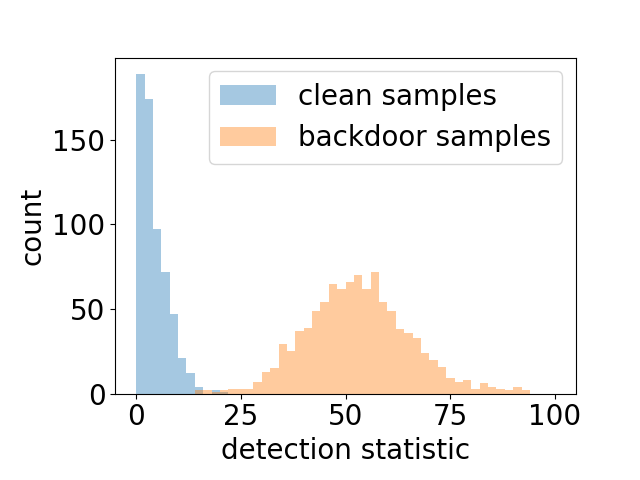}}
		\subcaption*{BadNet}
		\centerline{\includegraphics[width=1.0\linewidth]{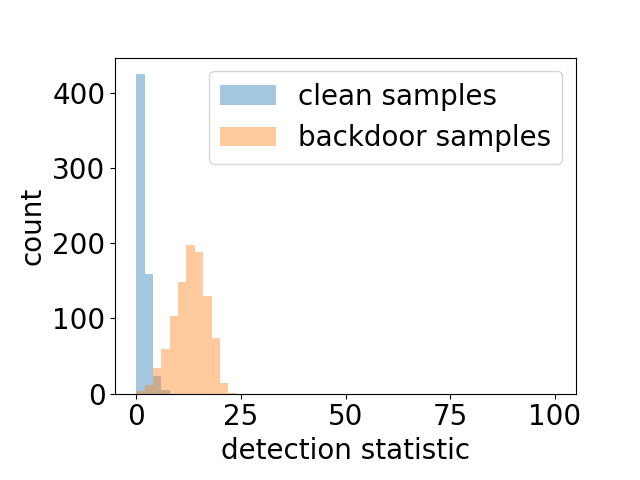}}
		\subcaption*{Input-aware}
	\end{minipage}
	\begin{minipage}[b]{.30\linewidth}
		\centering
		\centerline{\includegraphics[width=1.0\linewidth]{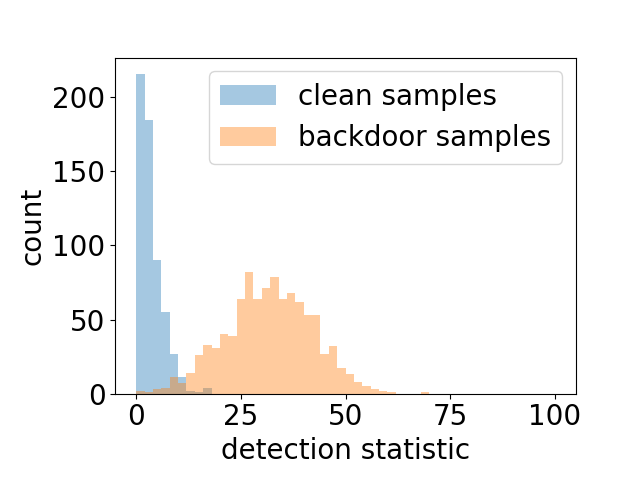}}
		\subcaption*{blend}
		\centerline{\includegraphics[width=1.0\linewidth]{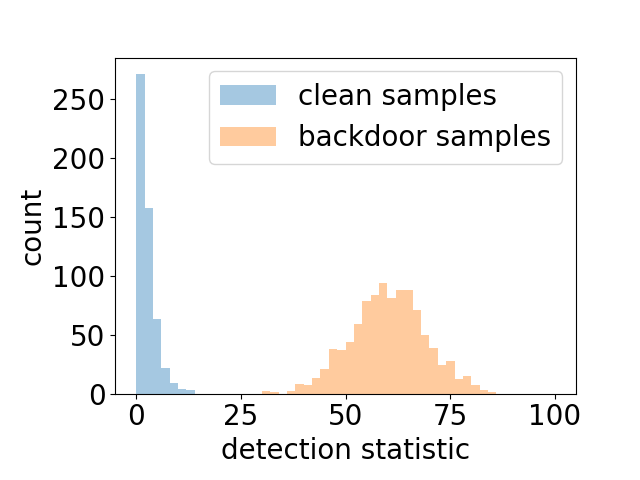}}
		\subcaption*{Clean-label}
	\end{minipage}
	\caption{MMDF detection statistics for clean samples and backdoor samples under different attacks.}
	\label{fig:distribution}
\end{figure*}
\begin{figure*}[ht!]

	\centering
	\begin{minipage}[b]{.30\linewidth}
		\centering
		\centerline{\includegraphics[width=1.0\linewidth]{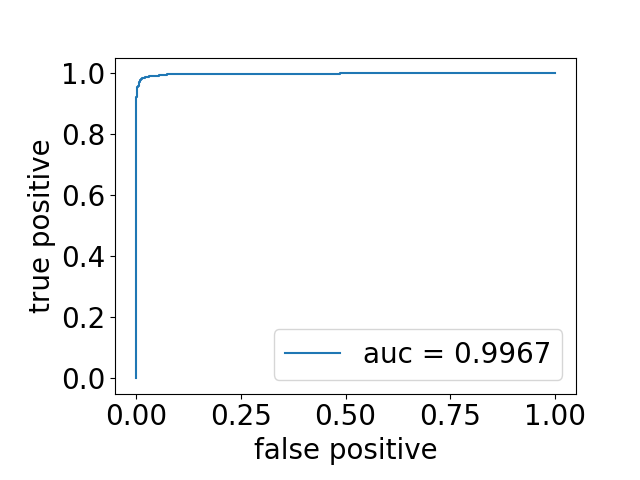}}
		\subcaption*{chessboard}
		\centerline{\includegraphics[width=1.0\linewidth]{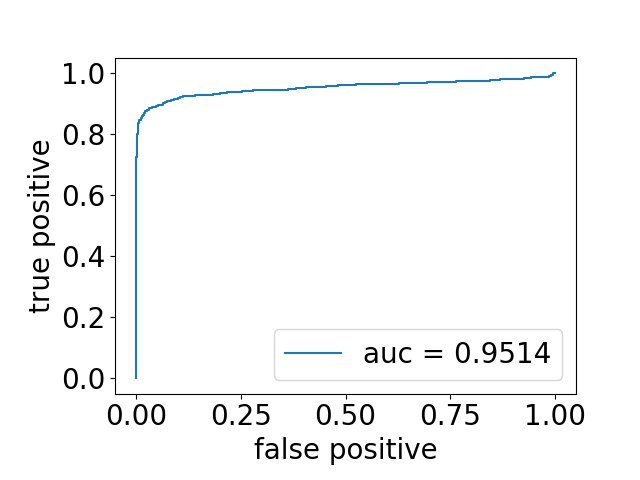}}
		\subcaption*{WaNet}
	\end{minipage}
	\begin{minipage}[b]{.30\linewidth}
		\centering
		\centerline{\includegraphics[width=1.0\linewidth]{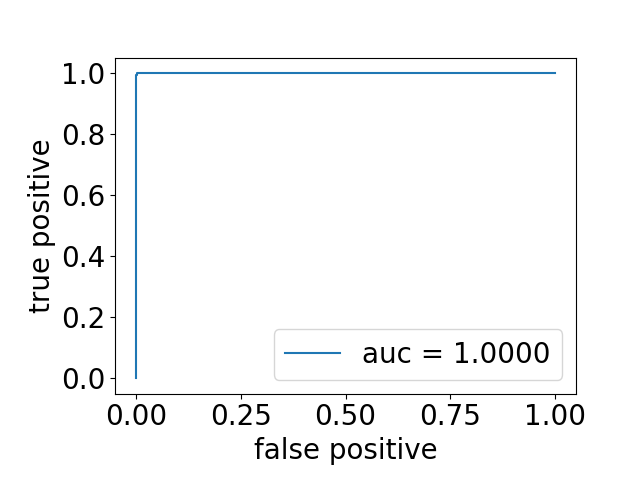}}
		\subcaption*{BadNet}
		\centerline{\includegraphics[width=1.0\linewidth]{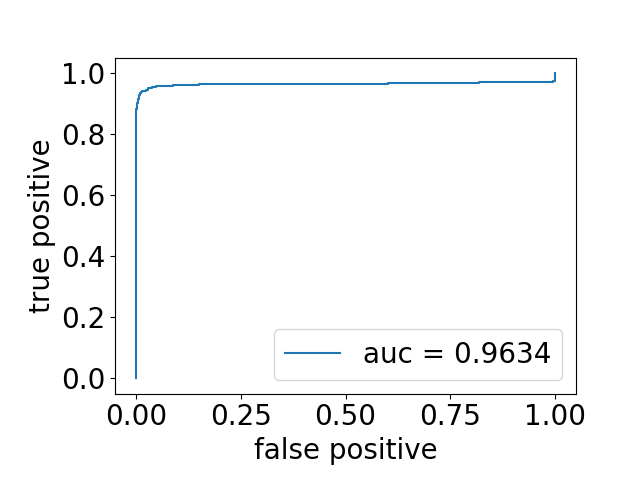}}
		\subcaption*{Input-aware}
	\end{minipage}
	\begin{minipage}[b]{.30\linewidth}
		\centering
		\centerline{\includegraphics[width=1.0\linewidth]{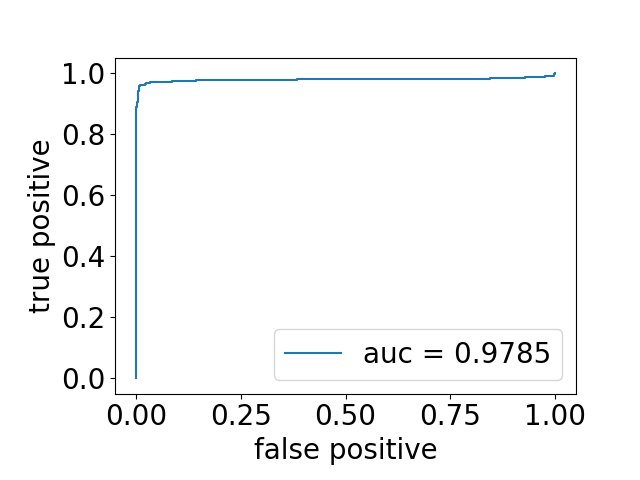}}
		\subcaption*{blend}
		\centerline{\includegraphics[width=1.0\linewidth]{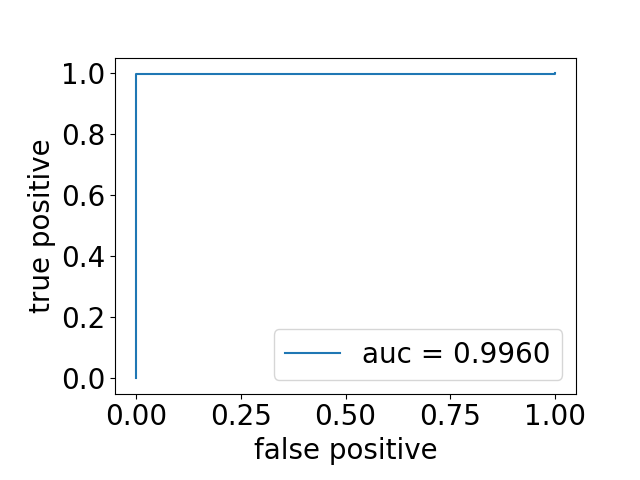}}
		\subcaption*{Clean-label}
	\end{minipage}
	\caption{Detection ROC curves for MMDF under different attacks.}
	\label{fig:roc}
\end{figure*}
In Eq. \eqref{eq:stat}, we proposed a detection statistic that can be used to detect test samples with backdoor triggers. Here we show the distributions of clean and backdoor-triggered samples. For each type of attack considered in Tab.  \ref{tab:mitigation}, 
we randomly selected the samples to poison, and also 
randomly chose 1000 clean samples and 1000 backdoor samples for evaluating 
test-time detection performance. The statistics for those samples are shown in Fig. \ref{fig:distribution}. For BadNet and Clean-label attacks (which both use a 3$\times$3 replaceable patch that is perceptible), backdoor samples and clean samples are clearly discriminated by the statistic. However, for other attacks, there is overlap between the statistic values for the backdoor and clean samples. We then produced the detection ROC curves under different attacks by changing the detection threshold. The ROC curves in Fig. \ref{fig:roc} and corresponding areas under curves (AUCs) indicate that our detection statistic is effective under all of the attacks.  

\end{document}